\documentclass[letterpaper]{article} 
\usepackage{aaai2026}  
\usepackage{times}  
\usepackage{helvet}  
\usepackage{courier}  
\usepackage[hyphens]{url}  
\usepackage{graphicx} 
\usepackage{natbib}  
\usepackage{caption} 
\usepackage{algorithm}
\usepackage{algorithmic}
\usepackage{amsmath}
\usepackage{amssymb}
\usepackage{booktabs}
\usepackage[table]{xcolor}
\usepackage{multirow}
\usepackage{newfloat}
\usepackage{listings}
\DeclareCaptionStyle{ruled}{labelfont=normalfont,labelsep=colon,strut=off} 
\floatstyle{ruled}
\newfloat{listing}{tb}{lst}{}
\floatname{listing}{Listing}
\title{ReaSon: Reinforced Causal Search with Information Bottleneck \\ for Video Understanding}

\author {
    Yuan Zhou\textsuperscript{\rm 1}\thanks{Corresponding author.},
    Litao Hua\textsuperscript{\rm 1},
    Shilong Jin\textsuperscript{\rm 1},
    Wentao Huang\textsuperscript{\rm 1},
    Haoran Duan\textsuperscript{\rm 2}
}
\affiliations {
    \textsuperscript{\rm 1}Nanjing University of Information Science and Technology\\
    \textsuperscript{\rm 2}Tsinghua University\\
    \{zhouyuan, 202412621441, 202312490769, 202412491490\}@nuist.edu.cn, haoran.duan@ieee.org
}

\begin{document}

\maketitle

\begin{abstract}
Keyframe selection has become essential for video understanding with vision-language models (VLMs) due to limited input tokens and the temporal sparsity of relevant information across video frames. Video understanding often relies on effective keyframes that are not only informative but also causally decisive. To this end, we propose \textit{\textbf{Re}inforced C\textbf{a}usal \textbf{S}earch with Inf\textbf{o}rmation Bottle\textbf{n}eck (ReaSon)}, a framework that formulates keyframe selection as an optimization problem with the help of a novel Causal Information Bottleneck (CIB), which explicitly defines keyframes as those satisfying both predictive sufficiency and causal necessity. Specifically, ReaSon employs a learnable policy network to select keyframes from a visually relevant pool of candidate frames to capture predictive sufficiency, and then assesses causal necessity via counterfactual interventions. Finally, a composite reward aligned with the CIB principle is designed to guide the selection policy through reinforcement learning. Extensive experiments on NExT-QA, EgoSchema, and Video-MME demonstrate that \textit{ReaSon} consistently outperforms existing state-of-the-art methods under limited-frame settings, validating its effectiveness and generalization ability. Code is available at: https://github.com/robin-hlt/AAAI26-ReaSon.
\end{abstract}

\section{Introduction}
\label{sec:intro}
Recent advances in video understanding have been greatly driven by the rise of vision-language models (VLMs)~\cite{Tang2025SurveyVLM,Nguyen2024VLUSurvey,feng2024efficient}. However, these models are severely constrained by input token budgets and suffer from the intrinsic redundancy of videos, where informative evidence is often sparsely distributed~\cite{wang2024videoagent,Ye2025TemporalSearch,Cao2025MASR}. As a result, recent studies have increasingly focused on developing frame selection strategies that extract a subset of keyframes~\cite{Ye2025TemporalSearch,VideoTree,Ma2025DrVideo,Akeys2025} to improve both computational efficiency and reasoning accuracy for video understanding. Yet a fundamental question remains unresolved: \textit{What defines a keyframe that is essential for video understanding?}

\begin{figure}[t]
\includegraphics[width=1.0\linewidth]{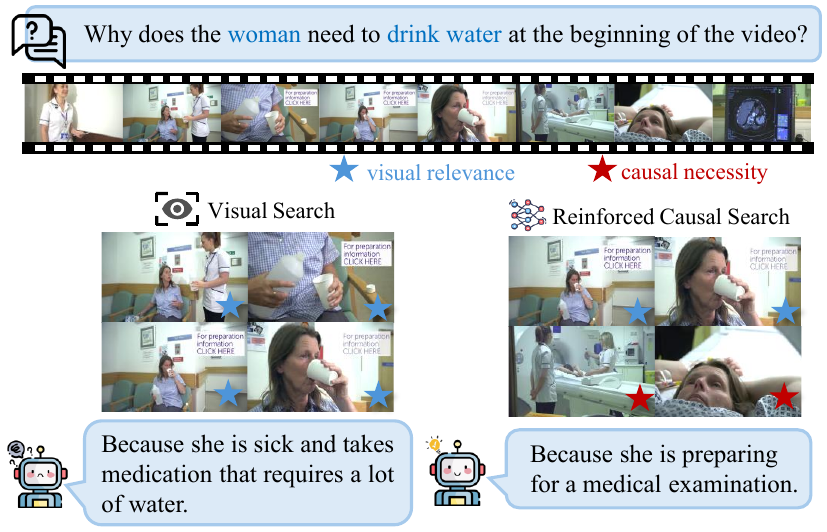}
\centering
    \caption{Illustration of limitations of visual relevance and the importance of causal necessity in keyframe selection. The visual search method selects visually relevant frames (blue stars) but misses causally decisive frames (red stars). In contrast, reinforced causal search captures causally necessary frames, leading to more accurate answers.}
    \label{fig:motivation}
\end{figure}

Most current methods~\cite{Ye2025TemporalSearch,Guo2025LogicInFrames,wang2024videoagent} define keyframes as an informative and compact subset of frames. These approaches typically select frames that are visually or semantically aligned with the question or answer, treating such correlation as a proxy for informativeness while restricting the number of frames to achieve compactness~\cite{Guo2025LogicInFrames,Ye2025TemporalSearch,wang2024videoagent}. This process implicitly adheres to the Information Bottleneck (IB) principle~\cite{Tishby2000IB}, which aims to preserve task-relevant information while discarding redundancy under a compression constraint. However, a high visual or semantic correlation does not guarantee decisive evidence for VLMs reasoning, which results from a lack of causal dependencies. As illustrated in Fig.~\ref{fig:motivation}, frames that appear visually relevant may not always be necessary for the correct reasoning process, whereas causally decisive frames, such as prior causes and subsequent effects, are often overlooked.

\begin{figure*}[htbp]
    \centering
    \includegraphics[width=0.9\linewidth]{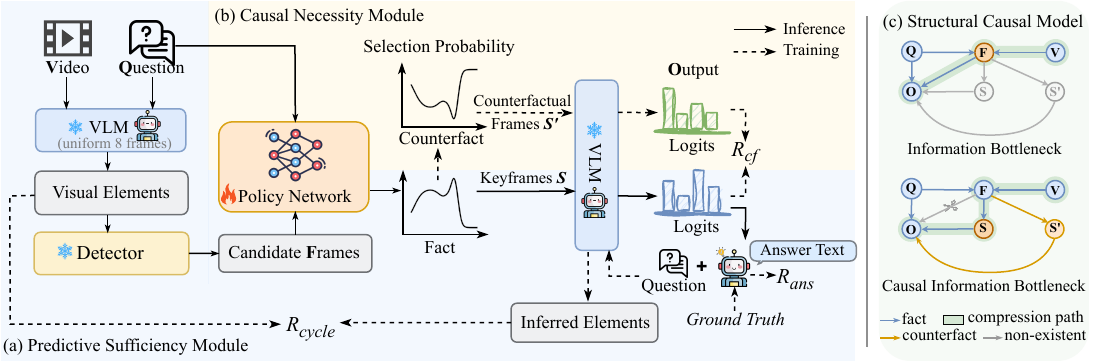}
    \caption{Framework of proposed ReaSon. (a) and (b) illustrate the predictive sufficiency and causal necessity modules, respectively, where a policy network learns to select keyframes based on CIB-aligned rewards. (c) shows the structural causal models. \(Q\) and \(V\) denote the question and the video, respectively. \(F\) and \(S\) represent selected frame subsets. \(S'\) is a counterfactual selection to assess causal necessity. \(O\) means the output. Bottleneck variables are highlighted with orange circles.}
    \label{fig:pipeline}
\end{figure*}

Motivated by this limitation, we revisit the concept of keyframes from a causal perspective~\cite{Yu2025CausalCoT}. Keyframes should meet two essential criteria: 1) predictive sufficiency, which ensures the selected subset supports accurate inference, and 2) causal necessity, which means no frame in the subset can be removed without impairing the output. To this end, we introduce the Causal Information Bottleneck (CIB) for keyframe selection, which extends the classic IB framework by incorporating an interventional term to capture predictive sufficiency and causal necessity. Furthermore, we present ReaSon, a reinforced causal search method that utilizes a learnable policy network to guide the keyframe selection process grounded in the CIB principle.

ReaSon consists of two core components: a predictive sufficiency module and a causal necessity module. The predictive sufficiency module first constructs a candidate pool by detecting question-relevant visual elements across the video. A learnable policy network is then leveraged to select a compact subset from the candidate pool. This two-stage process allows the model to initially localize potentially relevant regions via visual grounding and subsequently distill a subset of frames that are sufficient for reasoning, thus effectively reducing redundancy. In addition, the causal necessity module evaluates whether each selected frame is causally indispensable. Counterfactual interventions are constructed by altering the selected frame subset to generate a counterfactual input. The resulting distributional changes in the output of the VLM are measured to assess causal necessity, based on the assumption that substantial output shifts imply causal dependence. These measurements provide learning signals that guide the policy network to distinguish frames that merely correlate with the answer from those that are causally decisive. To implement the CIB-guided frame selection, we employ reinforcement learning to train a selection policy and design three rewards aligned with predictive sufficiency and causal necessity. 

Finally, we conduct extensive experiments to validate the effectiveness of ReaSon across diverse video understanding scenarios. ReaSon is evaluated on three representative benchmarks: NExT-QA\cite{Xiao2021NextQA} for causal, temporal, and descriptive reasoning; EgoSchema~\cite{Mangalam2023EgoSchema} for egocentric video understanding; and Video-MME~\cite{Fu2025VideoMME} for long-form video understanding. Our method is compared against recent non-selection-based methods and state-of-the-art frame selection approaches. Results show that ReaSon achieves the best accuracy of 81.4\% on NExT-QA validation set and 72.2\% on EgoSchema subset with only 8 frames. On Video-MME, ReaSon outperforms state-of-the-art methods, reaching improvements of 2.6\% (8 frames) and 2.3\% (32 frames). Ablation studies demonstrate that both the sufficiency and necessity modules make substantial contributions to performance. ReaSon consistently improves reasoning accuracy across different base models, highlighting its strong generalization ability.

In summary, our main contributions are as follows:
\begin{itemize}
\item We formally formulate keyframe selection as an information-theoretic optimization problem under the CIB, integrating predictive sufficiency and causal necessity.

\item We present ReaSon, a CIB-based causal search framework consisting of two dedicated modules that collaboratively identify sufficient and necessary keyframes.

\item Three CIB-aligned rewards are introduced to guide the selection policy via reinforcement learning, capturing answer correctness, semantic consistency, and causal necessity.

\item ReaSon outperforms existing frame selection and non-selection baselines under limited-frame settings, demonstrating strong effectiveness and generalization across diverse video understanding scenarios.
\end{itemize}

\section{Method}
\label{sec:method}
\subsection{Preliminary}
\label{sec:Preliminary}
Formally, let \( V \)\footnote{All uppercase letters denote random variables, and lowercase letters denote their instances unless stated otherwise.} and \( Q \) denote the input video and question as random variables and let \( F \) denote a subset of frames selected from the video \( V \). The goal of keyframe selection is to identify \( F \) that preserves the relevant information in \( V \) and \( Q \) for the output \( O \). As illustrated in Fig.~\ref{fig:pipeline}(c), this process can be described by a structural causal model (SCM)~\cite{pearl2009causality}. \( F \) serves as an intermediate variable that compresses the input \( V \), which is structurally aligned with the Information Bottleneck (IB) principle~\cite{Tishby2000IB}. Therefore, the objective of keyframe selection can be naturally formulated under the IB framework as:
\begin{equation}
\max \; \mathcal{I}(F; O) \quad \text{s.t.} \; \mathcal{I}(V, Q; F) \leq \beta,
\label{eq:ib}
\end{equation}
where \(\beta\) controls the allowed amount of information retained from the input and \( \mathcal{I} \) denotes mutual information. This objective encourages the selected frames to be informative for the output and as compressed as possible with respect to the input.

In practice, most existing frame selection methods follow IB principle by selecting a compact subset of frames that preserve high visual relevance to the question and answer. These approaches approximate the objective of maximizing \(\mathcal{I}(F; O)\) by prioritizing visually aligned frames under a fixed input budget. For example, VideoAgent~\cite{wang2024videoagent} predicts intermediate event descriptions using a large language model and retrieves frames with high image-text similarity to those events, while T*~\cite{Ye2025TemporalSearch} employs object detectors to locate frames containing entities mentioned in the question. However, by focusing solely on visual or semantic correlations, these methods may overlook decisive frames that are indispensable for correct reasoning.

\subsection{Causal Information Bottleneck}
\label{sec:cib}
To support reliable reasoning in video understanding, effective keyframes should satisfy two essential criteria:
\begin{itemize}
 \item \textbf{Predictive Sufficiency}: The selected frames must provide enough information to accurately answer the question, yielding outputs consistent with those derived from the full video;
 \item \textbf{Causal Necessity}: The frames should be a minimal subset with no redundancy, which means removing any frame in the subset would significantly affect the output.
\end{itemize}

To capture both sufficiency and causal necessity, we extend the Information Bottleneck to Causal Information Bottleneck with a causal perspective. The original bottleneck variable \( F \) lacks the capacity to represent selection as an intervenable decision, making it unsuitable for analyzing causal necessity. As illustrated in Fig.~\ref{fig:pipeline}(c), a new variable \( S \), which denotes the target keyframes, is introduced to replace \( F \) as the information bottleneck. This structural adjustment isolates the causal effect of selection and enables formal analysis of both predictive sufficiency and necessity. The Causal Information Bottleneck objective can be defined as follows:
\begin{equation}
\max \; \mathcal{I}(S; O) + \mathcal{I}_c(O; \text{do}(S)) \quad \text{s.t.} \; \mathcal{I}(V, Q; S) \leq \beta,
\label{eq:cib}
\end{equation}
where \(\mathrm{do}(S)\) denotes an intervention on \(S\), and \(\mathcal{I}_c\) represents mutual information defined under causal interventions. \( \mathcal{I}(S; O) \) encourages predictive sufficiency, ensuring the selected keyframes retain enough information to infer the correct output. The second term \( \mathcal{I}_c(O; \text{do}(S)) \) quantifies the influence of keyframes \(S\) under interventions to measure the causal necessity. The constraint \( \mathcal{I}(V, Q; S) \leq \beta \) limits the information capacity, preventing overly redundant selections.

\subsection{Reinforced Causal Search}
Building on the CIB, we propose \textbf{ReaSon}, a reinforced causal search approach that employs a learnable policy \( \pi_\theta(S \mid F, Q) \) to select keyframes \(S\). As illustrated in Fig.~\ref{fig:pipeline}, ReaSon comprises two dedicated modules targeting predictive sufficiency and causal necessity, respectively. However, mutual information terms in the CIB are intractable to compute or differentiate in practice. Therefore, in this section, tractable approximations of the CIB objective are derived in detail and are connected to three rewards. These distinct rewards aligned with the CIB objective jointly guide policy learning via reinforcement learning.

\subsubsection{Predictive Sufficiency Module}
\label{sec:Predictive Sufficiency Module}
This module focuses on optimizing the first term of the CIB objective, \( \mathcal{I}(S; O) \). \( \mathcal{I}(S; O) \) can be expressed as:
\begin{equation}
\mathcal{I}(S; O) = \mathbb{E}_{p(s, o)} \left[ \log \frac{p(o \mid s)}{p(o)} \right].
\label{eq:first-term}
\end{equation}

Directly computing Eq.~\eqref{eq:first-term} is intractable in general cases. To derive a tractable surrogate, we introduce a variational distribution \( q_\phi(o \mid s) \) to approximate the true posterior \( p(o \mid s) \)~\cite{alemi2017VIB,kingma2013auto}, and rewrite Eq.~\eqref{eq:first-term} as:
\begin{equation}
\begin{aligned}
\mathcal{I}(S; O) 
&= \mathbb{E}_{p(s, o)} \left[ \log \frac{q_\phi(o \mid s)}{p(o)} + \log \frac{p(o \mid s)}{q_\phi(o \mid s)} \right] \\
&= \mathbb{E}_{p(s, o)} \left[ \log \frac{q_\phi(o \mid s)}{p(o)} \right] 
\; \\
&+\; \mathbb{E}_{p(s)} \left[ D_\mathrm{KL}(p(o \mid s) \,\|\, q_\phi(o \mid s)) \right] \\
&\geq \mathbb{E}_{p(s, o)} \left[ \log \frac{q_\phi(o \mid s)}{p(o)} \right].
\end{aligned}
\label{eq:mi-vlb}
\end{equation}

Furthermore, ignoring the marginal term \( p(o) \), which is independent of \( s \), yields a looser but tractable approximation:
\begin{equation}
\mathcal{I}(S; O) \gtrsim \mathbb{E}_{p(s,o)}[\log q_{\phi}(o \mid s)] \triangleq J_1(s).
\label{eq:lower_bound}
\end{equation}

The resulting surrogate objective \(J_1(s)\) provides a tractable approximation of predictive sufficiency, which serves as a foundation for the reward design in practice. As shown in Fig.~\ref{fig:pipeline}(a), given a specific video \(v\) and a question \(q\), we first construct a candidate frame pool \(f = \{f_1, f_2, \dots, f_M\} \subseteq v\) using a heuristic-based pre-selection strategy inspired by T*~\cite{Ye2025TemporalSearch}, which filters out visually irrelevant frames. \(M\) denotes the size of the candidate frame pool. Uniformly sampled frames are fed into a frozen VLM to extract target visual elements \(E_q \). The visual elements are then matched against all frames using an open-vocabulary detector to form the candidate pool. Then, a policy network \(\pi_\theta(S \mid F=f, Q=q)\) is introduced to model the distribution over possible frame subsets conditioned on the candidate pool \(f\) and the question \(q\). The policy network assigns a selection probability to each frame in the candidate pool \(f\), and a keyframe subset \(s \sim \pi_\theta(S \mid f, q)\) is then sampled via a multinomial process, subject to a cardinality constraint \(|s| \leq K\). \(K\) denotes the maximum number of selected frames. This acts as a practical proxy to enforce the constraint \( \mathcal{I}(V, Q; S) \leq \beta \). The selected subset \(s\) with the question \(q\) is passed into the frozen VLM, which serves as an implementation of the variational distribution \( q_\phi(o \mid s) \), to generate an answer. A binary reward is computed by comparing the generated answer with the ground truth:
\begin{equation}
R_{\text{ans}} = \mathbb{I}[\mathrm{VLM}(s, q) = \text{gt}],
\end{equation}
where \( \mathrm{VLM}(s, q) \) denotes the generated textual output and \( \text{gt} \) is the ground truth. This reward serves as an approximation of the mutual information objective \( \mathcal{I}(S; O) \), encouraging the policy to select frames that lead to a correct answer.

\begin{table*}[t]
\centering
\small
\setlength{\tabcolsep}{6pt}
\renewcommand{\arraystretch}{1.1}
\begin{tabular}{l l c c c c c c}
\toprule
\multirow{2}{*}{\textbf{Method}} & 
\multirow{2}{*}{\textbf{VLM}} & 
\multirow{2}{*}{\textbf{Mean Frames}} & 
\multicolumn{4}{c}{\textbf{NExT-QA}} & 
\multirow{2}{*}{\textbf{EgoSchema}} \\
\cline{4-7}
& & & \textbf{Tem} & \textbf{Cau} & \textbf{Des} & \textbf{Avg} & \\
\midrule
\multicolumn{8}{c}{\textit{Non-selection Methods}} \\
\midrule
MVU \cite{rana2024mvu}       & Mistral-13B            & 16   & 55.4 & 48.1 & 64.1 & 55.2 & 60.3 \\
LangRepo \cite{kahatapitiya2024language}      & Mistral-8$\times$7B    & 180  & 51.4 & 64.4 & 69.1 & 60.9 & 66.2 \\
VideoChat2 \cite{li2024mvbench} & GPT-4                  & 16   & 57.4 & 61.9 & 69.9 & 61.7 & 54.4 \\
LLoVi \cite{zhang2024simple}      & GPT-4                  & --   & 61.0 & 69.5 & 75.6 & 67.7 & 61.2 \\
VideoINSTA \cite{liao2024videoinsta} & GPT-4               & 90   & -- & -- & -- & 72.3 & 65.0 \\
\midrule
\multicolumn{8}{c}{\textit{Frame Selection Methods}} \\
\midrule
VideoAgent \cite{wang2024videoagent} & GPT-4                 & 8.4   & 64.5 & 72.7 & 81.1 & 71.3 & 60.2 \\
VideoAgent \cite{Fan2024VideoAgent} & GPT-4                 & --   & 60.0 & 76.0 & 76.5 & 70.8 & 62.8 \\
LVNet \cite{Park2024TooManyFrames}      & GPT-4o                & 12    & 65.5 & 75.0 & 81.5 & 72.9 & 68.2 \\
DrVideo \cite{Ma2025DrVideo}    & GPT-4                 & 0.5fps& --   & --   & --   & --   & 66.4 \\
VideoTree \cite{VideoTree}  & GPT-4                 & 63.2  & 70.6 & 76.5 & 83.9 & 75.6 & 66.2 \\
AKEYS \cite{Akeys2025}          & GPT-4o                & 26.7  & 72.9 & 79.0 & 86.1 & 78.1 & 68.6 \\
T* \cite{Ye2025TemporalSearch}         & LLaVA-OneVision-7B    & 8     & --   & --   & --   & \underline{80.4} & 66.6 \\
\midrule
\multicolumn{8}{c}{\textit{Ours}} \\
\midrule
ReaSon & Qwen2.5-VL-7B      & 8 & \underline{76.4}   & \underline{81.0}   & \underline{86.6}   & \underline{80.4}   & 68.0 \\
ReaSon & LLaVA-Video-7B     & 8 & \textbf{77.3} & \textbf{82.1} & \textbf{87.4} & \textbf{81.4} & \underline{69.0} \\
ReaSon & GPT-4o             & 8 & 70.6 & 80.2 & 83.6 & 77.6 & \textbf{72.2} \\
\bottomrule
\end{tabular}
\caption{Comparison of ReaSon with existing state-of-the-art methods on NExT-QA and EgoSchema. We adopt accuracy (\%) as the metric. Results of baseline methods are directly cited from their respective publications. The best result is highlighted in bold, and the second-best is marked with underline.}
\label{tab:videoqa-comparison-main}
\end{table*}

To further reinforce predictive sufficiency, we introduce a cycle consistency reward that encourages semantic alignment throughout the reasoning process. After producing the final answer from the selected keyframes, the predicted answer is concatenated with the original question and sent back to the VLM to infer a set of visual elements, denoted as \(E_a\). Notably, the video frames are not accessible during this stage. The answer-based elements \(E_a\) are compared with the previously extracted target elements \(E_q\) to assess whether the reasoning process completes a semantic cycle: from visual input to answer reasoning and back to visual attribution. The cycle consistency reward is defined as:
\begin{equation}
R_{\text{cycle}} = \text{IoU}(E_q, E_a).
\label{eq:cycle_consistency_reward}
\end{equation}

A strong alignment between \(E_q\) and \(E_a\) indicates that the selected keyframes successfully preserve the semantic cues required to answer the question. This reward complements the answer reward \(R_{\text{ans}}\), providing additional guidance from the perspective of semantic consistency.

\subsubsection{Causal Necessity Module}
\label{sec:Causal Necessity Module}
This module is designed to optimize the second term \(\mathcal{I}_c(O; \text{do}(S))\) in the CIB, which can be expanded as follows:
\begin{equation}
\mathcal{I}_c(O; \text{do}(S)) = \mathbb{E}_{s \sim \pi_\theta} \left[ D_{\mathrm{KL}}\left(p(o \mid \text{do}(s)) \,\|\, p(o)\right) \right].
\label{eq:causal_MI_expand}
\end{equation}

According to the Causal Markov condition~\cite{pearl2009causality}, once the direct causes of a variable are fixed, the variable is conditionally independent of all other variables that are not its effects or direct causes. In our case, the output \(O\) is directly determined by the selected frames \(S\). Under the assumption that the structural mechanism from \(S\) to \(O\) remains unchanged under intervention~\cite{pearl2009causality}, we can treat \(do(S=s)\) as equivalent to conditioning on \(S=s\), and rewrite Eq.~\eqref{eq:causal_MI_expand} as:
\begin{equation}
\mathcal{I}_c(O; \text{do}(S)) = \mathbb{E}_{s \sim \pi_\theta} \left[ D_{\mathrm{KL}}\left(p(o \mid s) \,\|\, p(o)\right) \right].
\label{eq:causal_MI_rewrite}
\end{equation}

To address the intractability of computing the marginal distribution \(p(o)\) and assess the causal necessity of selected frames, a counterfactual selection strategy is introduced to approximate \(p(o)\) by inverting the original selection policy, denoted as \(\tilde{\pi}\). The counterfactual strategy can be defined as:
\begin{equation}
\tilde{\pi}(f_i) = \frac{1 - \pi_\theta(f_i)}{\sum_{j=1}^{M} \left(1 - \pi_\theta(f_j)\right)}, \; i,j \in [1, \dots, M],
\label{eq:inverted_policy}
\end{equation}
where \(\pi_\theta(f_i)\) and \(\tilde{\pi}(f_i)\) denote the original and the counterfactual selection probability assigned to the \(i\)-th frame, respectively. As shown in Fig.~\ref{fig:pipeline}(b), a counterfactual subset \(s'\) is sampled according to \(\tilde{\pi}\), which serves as a contrastive sample. Therefore, \(p(o)\) is computed as:
\begin{equation}
p(o) = \mathbb{E}_{s' \sim \tilde{\pi}} \left[ p(o \mid s') \right].
\label{eq:approx_po}
\end{equation}

Finally, substitute this approximation into Eq.~\eqref{eq:causal_MI_rewrite} and obtain the following objective:
\begin{equation}
\mathcal{I}_c(O; \text{do}(S)) = \mathbb{E}_{s \sim \pi_\theta} \left[ 
    D_{\mathrm{KL}}\left( 
        p(o \mid s) \,\middle\|\, 
        \mathbb{E}_{s' \sim \tilde{\pi}} [p(o \mid s')] 
    \right) 
\right].
\label{eq:causal_MI_surrogate}
\end{equation}

To reduce the computational cost during reinforcement learning, we approximate the expectation over counterfactual selections using a single Monte Carlo sample~\cite{alemi2017VIB} and rewrite Eq.~\eqref{eq:causal_MI_surrogate} as:
\begin{align}
\mathcal{I}_c(O; \text{do}(S)) 
&\approx \mathbb{E}_{s \sim \pi_\theta,\, s' \sim \tilde{\pi}} 
\left[ D_{\mathrm{KL}}(p(o \mid s) \,\|\, p(o \mid s')) \right] \notag \\
&\triangleq J_2(s, s').
\label{eq:contrastive_lower_bound}
\end{align}

In practice, to make the optimization of the surrogate objective \(J_2(s, s')\) feasible, a counterfactual reward is defined to guide the selection policy. Specifically, \( o \) and \( o' \) denote the logits outputs from the VLM given inputs \( (s, q) \) and \( (s', q) \), respectively. The reward can be formulated as:
\begin{equation}
R_{\text{cf}} = D_{\text{KL}}\left(\text{softmax}(o) \,\|\, \text{softmax}(o')\right).
\label{eq:cf_reward}
\end{equation}

This reward measures the effect of counterfactual interventions on the output to capture the causal necessity. Greater divergence implies that the absence of keyframes leads to the significant changes.

\subsection{Optimization}
The CIB objective can be finally approximated as:
\begin{equation}
\max_{s \sim \pi_\theta} J(s, s')=J_1(s)+J_2(s,s') \quad \text{s.t.} \; |s| \leq K.
\label{eq:cib_optimization}
\end{equation}

We construct a composite reward function \(R\) aligned with \( J(s,s') \) as a practical proxy that enables implicit optimization of the CIB objective via reinforcement learning. The reward \( R \) is defined as a weighted combination of the three rewards introduced above:
\begin{equation}
R =  R_{\text{ans}} + \lambda_1 R_{\text{cycle}} + \lambda_2 R_{\text{cf}}.
\label{eq:total_reward}
\end{equation}

\begin{table*}[t]
\centering
\small
\setlength{\tabcolsep}{6pt}
\renewcommand{\arraystretch}{1.1}
\begin{tabular}{l l c c c c c}
\toprule
\textbf{Method} & \textbf{VLM} & \textbf{Frames} & \textbf{Short} & \textbf{Medium} & \textbf{Long} & \textbf{Overall} \\
\midrule
Video-LLaVA~\cite{lin2023videollava}         &  Vicuna-7B v1.5    & 8   & 45.3 & 38.0 & 36.2 & 39.9 \\
LongVA~\cite{Zhang2024LongVA}              & Qwen2-7B               & 8   & 55.1 & 46.3 & 42.1 & 47.9 \\
GPT-4o              & GPT-4o                 & 8   & 55.7 & 54.3 & 51.4 & 53.8 \\
T*~\cite{Ye2025TemporalSearch}                  & GPT-4o     & 8   & 56.4 & \textbf{57.3}   & \textbf{56.4}   & 56.5   \\
\rowcolor{gray!10} ReaSon                & LLaVA-Video-7B         & 8   &  63.7 & 48.7 & 47.4 & 53.3 \\
\rowcolor{gray!10} Reason                & GPT-4o         & 8   & \textbf{65.9} & 57.1 & 54.4 & \textbf{59.1} \\
\midrule
LongVA              & Qwen2-7B               & 32  & 61.1 & 48.8 & 45.4 & 51.8 \\
LLaVA-NeXT-Video~\cite{Zhang2024LLaVANext}    & LLaVA-NeXT-Video-34B   & 32  & 61.7 & 50.1 & 44.3 & 52.0 \\
GPT-4o              & GPT-4o                 & 32  & 68.3 & 60.7 & 56.3 & 61.8 \\
T*                  & GPT-4o     & 32  & 69.5   & 63.5   & \textbf{59.3}   & 64.1 \\
\rowcolor{gray!10} ReaSon                & LLaVA-Video-7B        & 32  & 69.2 & 55.0 & 49.3 & 57.9 \\
\rowcolor{gray!10} ReaSon                & GPT-4o         & 32  & \textbf{76.8} & \textbf{64.2} & 58.2 & \textbf{66.4} \\
\midrule
Video-XL~\cite{Shu2025VideoXL}            & Qwen2-7B               & 128 & 64.0 & 53.2 & 49.2 & 55.5 \\
VideoChat-Flash~\cite{li2024videochat} & Qwen2-7B & 512 & -- & -- & 55.4 & 65.3 \\
VideoLLaMA 3~\cite{damonlpsg2025videollama3} & Qwen2.5-7B & 180 & 80.1 & 63.7 & 54.9 & 66.2\\
\midrule
\textcolor{gray}{Gemini 1.5 Pro}      & \textcolor{gray}{Gemini 1.5 Pro}         & \textcolor{gray}{1/0.5 fps} & \textcolor{gray}{81.7} & \textcolor{gray}{74.3} & \textcolor{gray}{67.4} & \textcolor{gray}{75.0} \\
\textcolor{gray}{GPT-4o}              & \textcolor{gray}{GPT-4o}                 & \textcolor{gray}{384} & \textcolor{gray}{80.0} & \textcolor{gray}{70.5} & \textcolor{gray}{65.3} & \textcolor{gray}{71.9} \\
\textcolor{gray}{Qwen2-VL}            & \textcolor{gray}{Qwen2-VL-72B}           & \textcolor{gray}{768} & \textcolor{gray}{80.1} & \textcolor{gray}{71.3} & \textcolor{gray}{62.2} & \textcolor{gray}{71.2} \\
\textcolor{gray}{LLaVA-Video}         & \textcolor{gray}{LLaVA-Video-72B}        & \textcolor{gray}{64}  & \textcolor{gray}{81.4} & \textcolor{gray}{68.9} & \textcolor{gray}{61.5} & \textcolor{gray}{70.6} \\
\bottomrule
\end{tabular}
\caption{Comparison of different methods on Video-MME without subtitles. We report accuracy (\%) across three video duration categories: Short (\textless 2 minutes), Medium (4-15 minutes), and Long (30-60 minutes). All baseline results are reported as cited from their respective publications. Our method are highlighted with a gray background. Best results in each setting are shown in bold. Noting that methods displayed in gray utilize significantly more frames and proprietary large models that are not publicly available or reproducible, making direct comparisons challenging.}
\label{tab:video-mme}
\end{table*}

To train the policy \(\pi_{\theta}\), we employ a group-wise policy gradient method~\cite{Chu2025GPG}, which estimates gradients based on multiple sampled selections per training instance. Specifically, for each video and question, we sample \( G \) subsets of keyframes \(\{s_i\}_{i=1}^G \sim \pi_{\theta}\) by multinomial sampling, along with a counterfactual subset \(s' \sim \tilde{\pi}\) for comparison. The corresponding model outputs \(o_i\) and \(o'\) are obtained via VLM and the rewards \(R_i\) are computed as defined in Eq.~\eqref{eq:total_reward}. To stabilize learning, intra-group advantages \(\hat{A}_i\) are calculated by mean-centering rewards, as defined in Eq.~\eqref{eq:normalized_advantage}, which reduces gradient variance while avoiding the bias and instability introduced by standard deviation normalization.
\begin{equation}
\hat{A}_i = R_i - \frac{1}{G} \sum_{j=1}^G R_j.
\label{eq:normalized_advantage}
\end{equation}

Finally, the selection policy is updated via policy gradient:
\begin{equation}
\nabla_{\theta} \mathcal{L} = \frac{1}{G} \sum_{i=1}^G \hat{A}_i \cdot \nabla_{\theta} \log \pi_{\theta}(s_i \mid f, q).
\label{eq:gpg_gradient}
\end{equation}

\section{Experiments}

\subsection{Experimental Setup}

\subsubsection{Datasets} 
We train ReaSon on NExT-QA~\cite{Xiao2021NextQA} training set and evaluate its performance on NExT-QA validation set, EgoSchema subset~\cite{Mangalam2023EgoSchema} and Video-MME~\cite{Fu2025VideoMME}. NExT-QA consists of 5,440 videos, which is designed to test temporal and causal reasoning over short videos. EgoSchema contains 5000 egocentric three-minute videos paired with multiple-choice questions, but only provides public labels for a subset of 500 questions. Video-MME is a recent-proposed long video understanding dataset, with an average video duration of 44 minutes. These datasets cover different video types and reasoning styles, including causal, temporal, egocentric and long video understanding.

\subsubsection{Implementation Details}
All videos are sampled at 1 fps in our proposed method. We leverage YOLO-World~\cite{Cheng2024YOLOWorld} as the detector to match visual elements in predictive sufficiency module. BLIP~\cite{Li2022BLIP} is used to encode both video frames and questions as the input to the policy network. The policy network consists of a three-layer LSTM~\cite{Hochreiter1997LSTM} and an MLP~\cite{Rumelhart1986Backprop}. During training, we set the candidate frame pool size to 32 and select 8 keyframes (\(M=32\) and \(K=8\)), using LLaVA-Video-7B~\cite{Zhang2024VideoInstruction} as VLM. \( \lambda_1 \) and \( \lambda_2 \) are set to 0.5. The number of groups \(G\) is set to 4. For inference, we employ LLaVA-Video-7B, Qwen2.5-VL-7B~\cite{Qwen2025VL}, and GPT-4o~\cite{Hurst2024GPT4o} across all datasets. For NExT-QA and EgoSchema, the keyframe selection settings remain consistent with those used during training. For Video-MME, the candidate frame pool size is increased to 64, with 32 keyframes selected to accommodate long video understanding (\(M=64\) and \(K=32\)). All 8-frame experiments are conducted on an RTX 3090 GPU, while 32-frame inference is performed on an A100 GPU.

\subsection{Comparison with Existing Approaches}

Table~\ref{tab:videoqa-comparison-main} shows a comparison of existing state-of-the-art methods and ReaSon on NExT-QA. We compare our method with both non-selection and frame selection approaches. ReaSon achieves the highest overall accuracy on NExT-QA (81.4\% with LLaVA-Video-7B) and state-of-the-art performance on EgoSchema (72.2\% with GPT-4o). Compared to non-selection methods such as VideoINSTA using 90 frames, ReaSon outperforms VideoINSTA by 9.1\% and 7.2\% under 8 frames, highlighting the importance of keyframe selection. Among frame selection methods, ReaSon also achieves top performance with fewer or comparable frames. Compared to the previous SOTA method AKEYS, ReaSon improves accuracy by 3.3\% on NExT-QA and 3.6\% on EgoSchema. Notably, the largest gains are observed on temporal and causal questions, with an improvement of 4.4\% and 3.1\% over AKEYS, showing the advantage of our method in modeling causal necessity. Furthermore, under the same frame setting, ReaSon outperforms T* by a substantial 5.6\% on EgoSchema. The best performance on each dataset is obtained using different VLMs equipped with our method, which is expected given the variation in data distribution and the robustness of VLMs. Importantly, ReaSon achieves strong and stable performance across all VLMs, consistently ranking among the top-performing methods on both datasets.

\begin{table}[t]
\centering
\setlength{\tabcolsep}{2.5pt}
\renewcommand{\arraystretch}{1.2}
\begin{tabular}{l c c c c c}
\toprule
\multirow{2}{*}{\textbf{Method}} & \multicolumn{4}{c}{\textbf{NExT-QA}} & \multirow{2}{*}{\textbf{EgoSchema}} \\
\cline{2-5}
& \textbf{Tem} & \textbf{Cau} & \textbf{Des} & \textbf{Avg} & \\
\midrule
w/ $R_{\text{ans}}$  & 76.3 & 81.1 & 84.4 & 80.1 & 66.0 \\
w/ $R_{\text{ans}}$ + $R_{\text{cycle}}$   & 76.8 & 81.7 & 84.3 & 80.5 & 68.2 \\
w/ $R_{\text{ans}}$ + $R_{\text{cycle}}$ + $R_{\text{cf}}$ & 77.3 & 82.1 & 87.4 & 81.4 & 69.0 \\
\bottomrule
\end{tabular}
\caption{Ablation study evaluating the contribution of different reward components in \textbf{ReaSon} on NExT-QA and EgoSchema. Results are reported using \textbf{LLaVA-Video-7B} with 8-frame input.}
\label{tab:reward-ablation}
\end{table}

\begin{table}[t]
\centering
\setlength{\tabcolsep}{4pt}
\renewcommand{\arraystretch}{1.1}
\begin{tabular}{l c c c c c}
\toprule
\multirow{2}{*}{\textbf{VLM}} & \multicolumn{4}{c}{\textbf{NExT-QA}} & \multirow{2}{*}{\textbf{EgoSchema}} \\
\cline{2-5}
& \textbf{Tem} & \textbf{Cau} & \textbf{Des} & \textbf{Avg} & \\
\midrule
LLaVA-Video-7B       & 76.3 & 81.4 & 83.9 & 80.2 & 65.2 \\
\textit{+ReaSon}        & 77.3 & 82.1 & 87.4 & 81.4 & 69.0 \\
\midrule
Qwen2.5-VL-7B        & 75.9 & 80.6 & 86.0 & 79.9 & 65.8 \\
\textit{+ReaSon}        & 76.4 & 81.0 & 86.6 & 80.4 & 68.0 \\
\midrule
GPT-4o               & 64.4 & 75.2 & 76.6 & 72.0 & 70.0 \\
\textit{+ReaSon}        & 70.6 & 80.2 & 83.6 & 77.6 & 72.2 \\
\bottomrule
\end{tabular}
\caption{Evaluation of the effectiveness and generalization ability of \textbf{ReaSon} across different VLMs on NExT-QA and EgoSchema. \textbf{ReaSon} is conducted under 8-frame input.}
\label{tab:Vlm-ablation}
\end{table}

\begin{figure*}[t]
    \centering
    \includegraphics[width=0.9\linewidth]{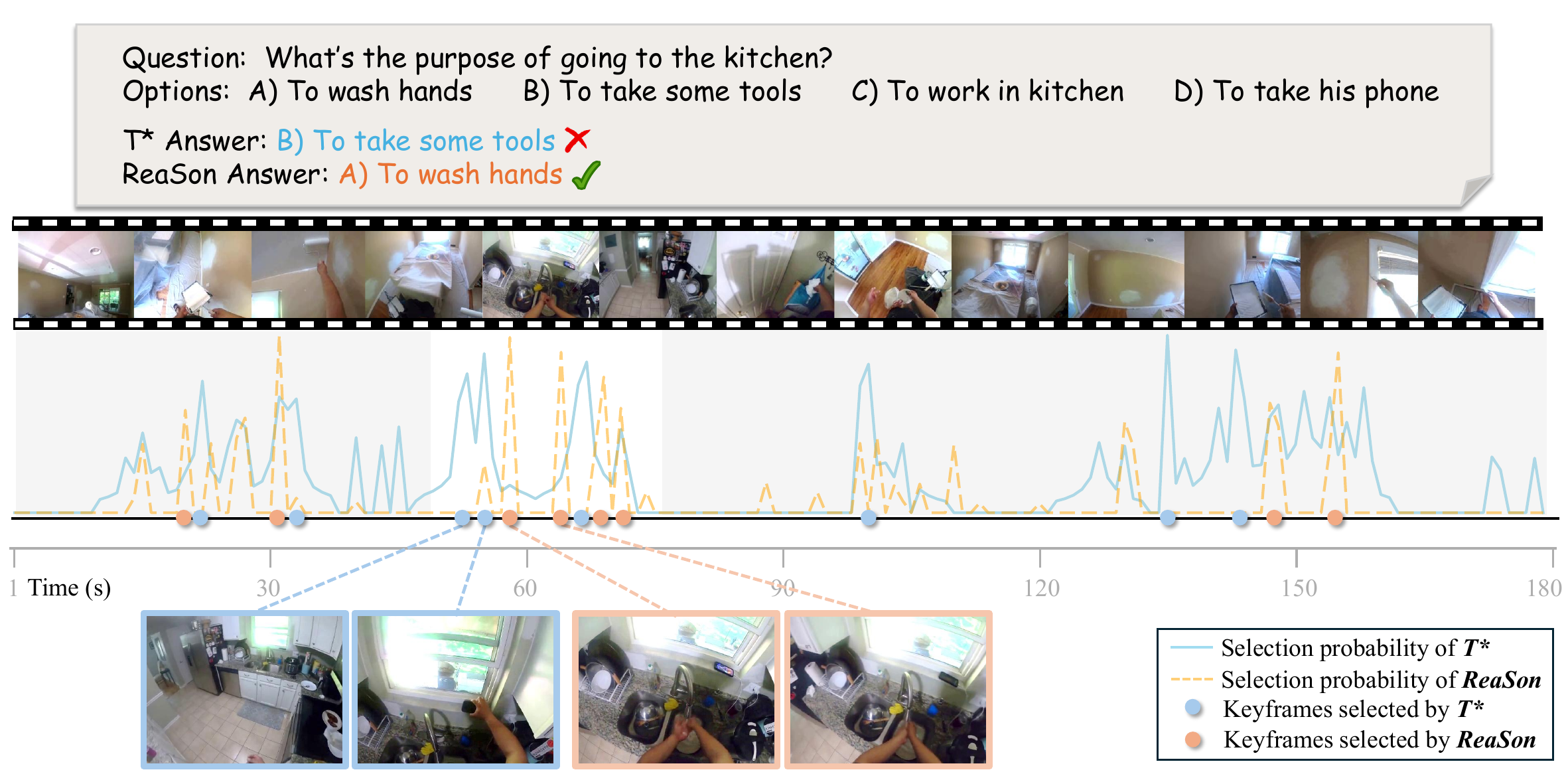}
    \caption{The visualization of frame selection results demonstrates the effectiveness of our approach compared to the previous state-of-the-art method \textit{T*}. Our approach pays less attention to irrelevant regions (in gray) and identifies more causal decisive keyframes.}
    \label{fig:visualization}
\end{figure*}

Additionally, we evaluate ReaSon’s performance on Video-MME for long video understanding. As shown in Table~\ref{tab:video-mme}, ReaSon with GPT-4o achieves the best overall accuracy of 59.1\% and 66.4\% among all methods under comparable configurations with 8 and 32 keyframes, respectively. This highlights the effectiveness of our selection strategy under limited frame budgets. For short videos, ReaSon demonstrates the most significant improvement. Under the 8-frame setting, ReaSon boosts GPT-4o performance from 55.7\% to 65.9\% and outperforms T* by 9.5\%. With 32 frames, ReaSon exceeds GPT-4o and T* by 8.5\% and 7.3\%, respectively. For medium and long videos, ReaSon exhibits steady improvements. Using 8 frames, it improves GPT-4o performance by 2.8\% and 3.0\% on medium and long videos, respectively. When extended to 32 frames, the gains over vanilla GPT-4o reach 3.5\% for medium and 1.9\% for long videos. ReaSon surpasses T* by 0.7\% on medium videos but falls short on long videos. Despite certain limitations on long videos, ReaSon remains competitive across diverse video lengths, even compared to methods with significantly larger frame budgets such as Video-XL, VideoChat-Flash, and VideoLLaMA 3.

\subsection{Ablation Study}

We conduct an ablation study to assess the effectiveness of reward components in ReaSon. As shown in Table~\ref{tab:reward-ablation}, each reward component in ReaSon contributes to the overall performance. The cycle consistency reward $R_{\text{cycle}}$ improves sufficiency by aligning visual semantics, while the counterfactual reward $R_{\text{cf}}$ enhances necessity by encouraging the selection of frames that are causally necessary for reasoning. The complete reward combination achieves the best results on both datasets.

We further evaluate the effectiveness and generalization of ReaSon with different VLMs. Table~\ref{tab:Vlm-ablation} shows the results using LLaVA-Video-7B, Qwen2.5-VL-7B, and GPT-4o. Comparing each VLM with and without ReaSon, we observe consistent performance improvements across all settings, confirming the effectiveness of ReaSon as a plug-in module for enhancing video understanding. The largest overall gains are observed on Qwen2.5-VL-7B (+3.8\% on EgoSchema) and GPT-4o (+5.6\% on NExT-QA). While different VLMs show dataset-specific strengths and no single model performs best across all scenarios, ReaSon does not rely on any specific model and can enhance each VLM, highlighting its robustness and generalization.

\subsection{Visualization}
Fig.~\ref{fig:visualization} presents the visualization of frame selection results on a video sampled from EgoSchema, paired with a manually crafted question. Compared to the previous state-of-the-art method T*, our approach identifies more relevant and causally decisive frames and correctly answers the question. During the selection process, our method anchors regions that are highly relevant to the question and also attends to their preceding and succeeding context to capture potential causal cues. As a result, the overall frame selection probability exhibits three prominent peaks. While T* is also able to localize question-relevant regions, its frame selection tends to include unnecessary frames, leading to broader and less precise peaks. In contrast, our method pays less attention to irrelevant regions and maintains only minimal focus on unrelated regions to avoid missing useful information. Within the highly relevant regions, ReaSon effectively captures causal necessity and filters out frames that are correlated but not essential.

\section{Conclusion}
In this paper, we introduce ReaSon, a reinforced causal search framework for video understanding, grounded in a novel Causal Information Bottleneck. By modeling keyframe selection as an optimization of both predictive sufficiency and causal necessity, ReaSon captures not only visually relevant but also causally decisive information. Through a dual-module architecture and reinforcement learning guided by a composite reward, our method identifies compact keyframe subsets that support accurate reasoning. Extensive experiments across diverse video types and question categories demonstrate that ReaSon consistently outperforms strong baselines under limited-frame settings while maintaining strong generalization across different VLMs. We believe our framework offers a principled and extensible foundation for efficient video understanding. In future work, we plan to further improve frame selection for more challenging long video scenarios.

\bibliography{main}

\clearpage
\appendix
\twocolumn[
\begin{center}
{\Large \textbf{Supplementary Materials for\\
ReaSon: Reinforced Causal Search with Information Bottleneck\\
for Video Understanding}}
\end{center}
\vspace{5em}
]

\section{Related Work}
\subsection{Video Understanding with Vision-Language Models}
Recent advances in vision-language models (VLMs)~\cite{Tang2025SurveyVLM,Nguyen2024VLUSurvey}, such as GPT-4o, LLaVA-Video~\cite{lin2023videollava}, and Qwen-VL~\cite{Qwen2025VL}, have significantly improved the performance of video understanding tasks including video question answering, captioning, and reasoning. These models integrate powerful large language models with visual encoders or adapters, enabling precise perception and multimodal alignment. However, directly applying VLMs to video understanding remains challenging due to excessive frame redundancy and limited context length~\cite{wang2024videoagent,Park2024TooManyFrames}. Videos often exhibit substantial temporal and spatial redundancy, while only a small fraction of frames carry essential information for answering the question. Feeding entire videos into VLMs not only imposes a heavy computational burden, but also obscures critical evidence with irrelevant context.

\subsection{Keyframe Selection for Video Understanding}

To address the challenge mentioned above, recent works~\cite{Ma2025DrVideo,Ye2025TemporalSearch,Guo2025LogicInFrames,Cao2025MASR,VideoTree} have proposed selecting a subset of keyframes as input to the VLM. The goal is to preserve relevant frames while reducing redundancy, thereby improving reasoning efficiency under strict input constraints. Despite their different implementations, these methods share a common pursuit: to identify a sufficient set of frames for accurate understanding. Existing keyframe selection methods can be broadly grouped into two categories: static heuristic selection, which relies on rule-based filtering or semantic matching, and agent-based interactive search, which performs step-wise frame exploration via LLM-guided reasoning.

\paragraph{Static Heuristic Selection.}
Static heuristic methods select keyframes based on predefined signals such as visual relevance, semantic similarity, or heuristic alignment with the question, typically without requiring training or iterative reasoning. For instance, \textbf{T*}~\cite{Ye2025TemporalSearch} introduces a multi-stage pipeline combining object grounding and confidence-based filtering to select frames that visually anchor the answer. \textbf{LVNet}~\cite{Park2024TooManyFrames} performs hierarchical filtering through visual clustering, question keyword alignment, and VLM-based template matching to progressively select relevant frames. \textbf{BOLT}~\cite{Liu2025BOLT} transforms frame-query similarity scores into probabilistic weights and applies inverse transform sampling to balance diversity and relevance. \textbf{VideoTree}~\cite{VideoTree} builds an adaptive tree of video segments via clustering, and selects a reasoning path that semantically aligns with the question through hierarchical traversal. While these methods are efficient and easy to integrate into existing base VLMs, they remain fundamentally static and heuristic. Their selection criteria are typically predefined and fixed across inputs, lacking adaptability to task complexity or reasoning context.

\paragraph{Agent-based Interactive Search.}
In contrast, agent-based approaches treat keyframe selection as an iterative and adaptive reasoning process. These methods simulate active evidence acquisition, where a VLM agent incrementally decides what to observe, retrieve, or retain based on prior outputs. \textbf{VideoAgent}~\cite{Fan2024VideoAgent} employs an LLM as a planner to iteratively select frames, predict answers and assess confidence. Based on the LLM's self-reflection, it retrieves additional frames, which are then captioned and stored in a memory. This process enables the agent to progressively refine its understanding. \textbf{AKEYS}~\cite{Akeys2025} performs keyframe selection by constructing a semantic search tree over frame subsets, where an LLM acts as an agent to iteratively expand nodes based on answer relevance and path cost. \textbf{DrVideo}~\cite{Ma2025DrVideo} builds a dynamic textual memory by retrieving semantically relevant frames and iteratively searching for missing evidence via agent interaction, which enables the LLM to perform chain-of-thought inference over a question-aware video document. \textbf{MASR}~\cite{Cao2025MASR} combines hierarchical vision-language attention with a self-reflective selection, gradually refining frame subsets based on confidence and multimodal relevance. Compared to static heuristics, these methods offer greater adaptability, but often require multi-step LLM inference to guide frame selection, resulting in higher VLM inference cost.

In summary, existing approaches to keyframe selection have made notable progress in reducing redundancy and improving efficiency for VLM-based video understanding. Static methods emphasize simplicity and speed but struggle with adaptivity, while agent-based methods offer greater flexibility at the cost of inference overhead. However, both paradigms largely rely on empirical heuristics or architectural intuition, and lack a unified theoretical foundation to guide the selection process. Moreover, existing methods typically emphasize visual relevance or confidence generated by VLM, yet visually relevant frames or high confidence may not be causally decisive for accurate answers. This highlights the urgent need for a theoretically grounded and efficient framework to identify truly effective keyframes for video understanding.

\section{Notation}
Notations used in this paper are summarized in Table~\ref{tab:notation}.
\begin{table}[htbp]
\centering
\small
\begin{tabular}{ll}
\toprule
\textbf{Symbol} & \textbf{Meaning} \\
\midrule
$V$ \& $v$ & Video \\
$Q$ \& $q$ & Question\\
$F$ \& $f$ & \textbf{In IB}: selected frame subset \\
    & \textbf{In CIB}: candidate frame pool \\
$f_i \in f$ & A single frame in a candidate pool \\
$S$ \& $s$ & Keyframe subset \\
$s_i$ & The $i$-th group of keyframe subset \\
$\pi_\theta(S \mid F, Q)$ & Selection policy over random variables \\
$\pi_\theta(S \mid f, q)$ & Selection policy given a specific $f$ and $q$ \\
$\pi_\theta$ & Abbreviation of the selection policy\\
$\pi_\theta(f_i)$ & Frame-wise selection probability under $\pi_\theta$ \\
$\tilde{\pi}(f_i)$ & Counterfactual selection probability of $f_i$\\
$\tilde{\pi}$ & Abbreviation of counterfactual policy \\
$S'$ \& $s'$ & Counterfactual frame subset sampled from $\tilde{\pi}$ \\
$O$ & Logits output from the VLM \\
$o$ & An instance of logits output given $(s, q)$ \\
$o'$ & An instance of logits output given $(s', q)$ \\
$\mathrm{VLM}(\cdot, \cdot)$ & Textual answer generated by the VLM\\
$\mathcal{I}(\cdot; \cdot)$ & Mutual information between two variables \\
$R_{ans}$ & Answer reward function \\
$R_{cycle}$ & Cycle consistency reward function \\
$R_{cf}$ & Counterfactual reward function \\
$R$ & Composite reward function \\
$R_i$ & The $i$-th group reward \\
$\hat{A}_i$ & The $i$-th group reward advantage \\
$M$ & The size of the candidate pool \\
$K$ & The number of the selected keyframes \\
$G$ & The number of the groups \\
\bottomrule
\end{tabular}
\caption{Notation summary used throughout the paper. Uppercase letters represent random variables, and the corresponding lowercase letters represent an instance.}
\label{tab:notation}
\end{table}

\section{Implementation Details}
\subsection{Experiments Compute Resources}
All training and inference operations with 8 keyframes can be easily conducted on single NVIDIA RTX 3090 GPU (24GB). In contrast, experiments with 32 keyframes require at least 48GB of GPU memory, for which we use single NVIDIA A100 GPU. We set the temperature to 0 for all experiments using GPT-4o, LLaVA-Video and Qwen.

\subsection{Datasets}
In this paper, we utilize three open-source video question-answering datasets: NExT-QA~\cite{Xiao2021NextQA}, EgoSchema~\cite{Mangalam2023EgoSchema}, and Video-MME~\cite{Fu2025VideoMME}. These datasets are chosen for their diversity in question types and video lengths, enabling a comprehensive evaluation of our method. A detailed description of each dataset is provided below.

\paragraph{NExT-QA} NExT-QA is a large-scale benchmark for video question answering. It consists of 5,440 real-world videos and 52,044 manually annotated question-answer pairs. In this work, we use only the multiple-choice QA part, which contains 34,132 training and 4,996 validation examples. Questions are divided into three types: descriptive (23\%), causal (48\%), and temporal (29\%). Each question is paired with a short video segment and requires reasoning over fine-grained temporal dynamics, object interactions, and causal relationships. 

\paragraph{EgoSchema} EgoSchema is a diagnostic benchmark designed to evaluate long-form video understanding. It contains over 5,000 manually curated multiple-choice QA instances, spanning more than 250 hours of egocentric video. Each question is grounded in a three-minute video clip. EgoSchema emphasizes not only temporal reasoning but also abstract understanding, including tasks such as summarizing overarching behaviors, inferring intentions, and integrating temporally dispersed cues. Human performance on EgoSchema reaches 76\% accuracy in the unconstrained setting. Only 500 questions in this dataset have publicly available labels.

\paragraph{Video-MME} Video-MME is a large-scale benchmark for evaluating multi-modal large language models (MLLMs) on video understanding. It contains 900 videos across six domains, ranging from 11 seconds to 1 hour, with a total of 2,700 multiple-choice questions. The questions cover diverse reasoning types, with a focus on temporal and compositional understanding. Each video is accompanied by subtitles and audio, enabling multi-modal evaluation. In this paper, we conduct all evaluations without using subtitles or audio.

\subsection{Training Details}
The detailed training process of ReaSon is represented in Algorithm~\ref{alg:training}.
\begin{algorithm}[t]
\caption{Training process of Reinforced Causal Search}
\label{alg:training}
\begin{algorithmic}[1]
\REQUIRE Dataset $\{(v^n, q^n)\}_{n=1}^N$, VLM, open-vocabulary detector, selection policy $\pi_\theta$, $\lambda_1=\lambda_2=0.5$, $G=4$
\FOR{each video-question pair $(v, q)$}
    \STATE $E_q \gets \mathrm{VLM}(Uniform(v), q)$
    \STATE $f \gets \texttt{Detector}(E_q)$
    \FOR{$i = 1$ to $G$}
        \STATE $s_i \sim \pi_\theta(S \mid f, q)$ \hfill // Multinomial sampling
        \STATE $o_i \gets \texttt{VLM Logits}(s_i, q)$ \hfill // Obtain logits
        \STATE $a_i \gets \mathrm{VLM}(s_i, q)$ \hfill // Generate Answer
        \STATE $E_a \gets \mathrm{VLM}(a_i, q)$
        \STATE $s'_i \sim \tilde{\pi}$; $o'_i \gets \texttt{VLM Logits}(s'_i, q)$
        \STATE $R_\mathrm{ans} \gets \mathbb{I}[\arg\max a_i = \text{gt}]$
        \STATE $R_\mathrm{cycle} \gets \texttt{IoU}(E_q, E_a)$
        \STATE $R_\mathrm{cf} \gets D_{\mathrm{KL}}(o_i \parallel o'_i)$
        \STATE $R_i \gets R_\mathrm{ans} + \lambda_1 R_\mathrm{cycle} + \lambda_2 R_\mathrm{cf}$
    \ENDFOR
    \STATE $\hat{A}_i \gets R_i - \frac{1}{G} \sum_{j=1}^G R_j$
    \STATE $\nabla_\theta \gets \frac{1}{G} \sum_{i=1}^G \hat{A}_i \cdot \nabla_\theta \log \pi_\theta(s_i \mid f, q)$
    \STATE Update $\pi_\theta$ via policy gradient
\ENDFOR
\RETURN Selection policy $\pi_\theta$
\end{algorithmic}
\end{algorithm}

\paragraph{Detector} We use YOLO-World~\cite{Cheng2024YOLOWorld}, an open-vocabulary detector, to identify the frames with target elements. The detection confidence threshold is set to 0.7.

\paragraph{Policy Network} We utilize a frozen BLIP~\cite{Li2022BLIP} as an input encoder,  followed by a learnable 3-layer LSTM~\cite{Hochreiter1997LSTM} and a MLP layer~\cite{Rumelhart1986Backprop}.

\paragraph{Sampling Strategy} During training, we adopt multinomial sampling from the learned policy distribution to encourage exploration and gradient diversity, which allows the policy to observe a wide range of frame combinations and receive informative reward signals.

\paragraph{Optimization} We use the Adam optimizer with a learning rate of 1e-4.

\subsection{Inference Details}
The inference procedure of ReaSon is detailed in Algorithm~\ref{alg:inference}. For inference efficiency, ReaSon achieves 7.52s latency on average.
\begin{algorithm}[t]
\caption{Inference process of Reinforced Causal Search}
\label{alg:inference}
\begin{algorithmic}[1]
\REQUIRE Test video-question pair $(v, q)$, trained policy $\pi_\theta$, frame budget $K$, VLM, open-vocabulary detector
\STATE $E_q \gets \mathrm{VLM}(Uniform(v), q)$
\STATE $f \gets \texttt{Detector}(E_q)$ 
\STATE $s \gets \texttt{TopK}(f, q, \pi_\theta, K)$ \\ // Select top-$K$ highest-probability frames
\STATE $a \gets \mathrm{VLM}(s, q)$
\STATE \textbf{return} Keyframes $s$, Answer $a$
\end{algorithmic}
\end{algorithm}

\paragraph{Sampling Strategy} During inference, we adopt a Top-K sampling strategy to select frames based on the learned policy distribution, which differs from the training phase. The inference process focuses on selecting the most promising frames deterministically.

\paragraph{Frame Count Considerations} We evaluate our method under fixed frame budgets (8 or 32 frames). These settings reflect practical constraints in real-world applications, where models are often limited by input token budgets or hardware resources. The 8-frame setting simulates a low-resource scenario that emphasizes selection efficiency, while 32 frames provide a loose condition to test the scalability and consistency of our method. Importantly, our goal is not to compete purely on absolute frame counts, but to evaluate how well the policy network selects informative subsets under constrained budgets.

\subsection{Prompt Design}
\begin{figure}[htbp]
\centering
    \includegraphics[width=0.9\linewidth]{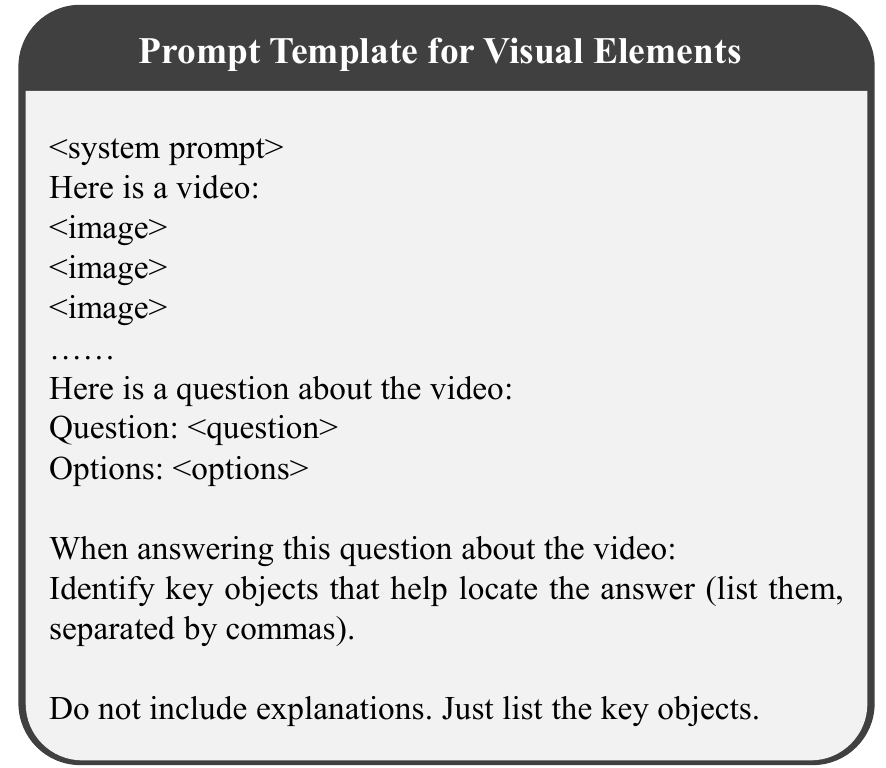}
    \caption{The prompt template for visual elements, where \texttt{\textless image\textgreater} represents a \texttt{PIL.Image} object for each frame, and other angle-bracketed tokens are strings.}
    \label{prompt1}
\end{figure}

\begin{figure}[htbp]
\centering
    \includegraphics[width=0.9\linewidth]{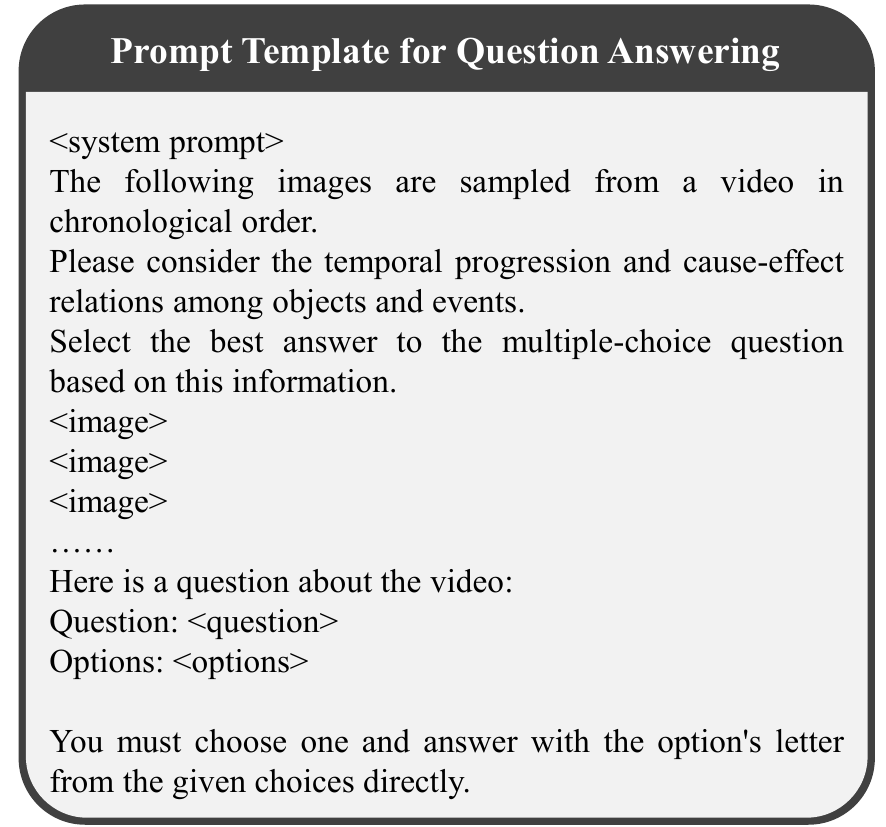}
    \caption{The prompt template for question answering.}
    \label{prompt2}
\end{figure}

\begin{figure}[htbp]
\centering
    \includegraphics[width=0.9\linewidth]{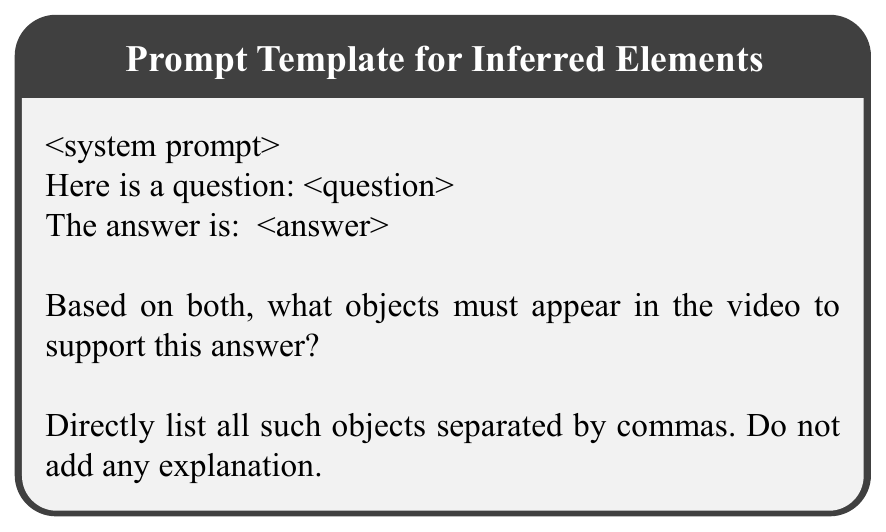}
    \caption{The prompt template for inferred elements.}
    \label{prompt3}
\end{figure}

In this section, all the prompts used in our method are included. Fig.~\ref{prompt1} shows the prompt template for visual element extraction in the predictive sufficiency module. The prompt in Fig.~\ref{prompt2} is used to answer questions based on the selected keyframes and question. The prompt shown in Fig.~\ref{prompt3} is used to extract visual elements from the answer and the question, facilitating the construction of a cycle consistency reward.

\begin{figure*}[htbp]
\centering
    \includegraphics[width=0.9\linewidth]{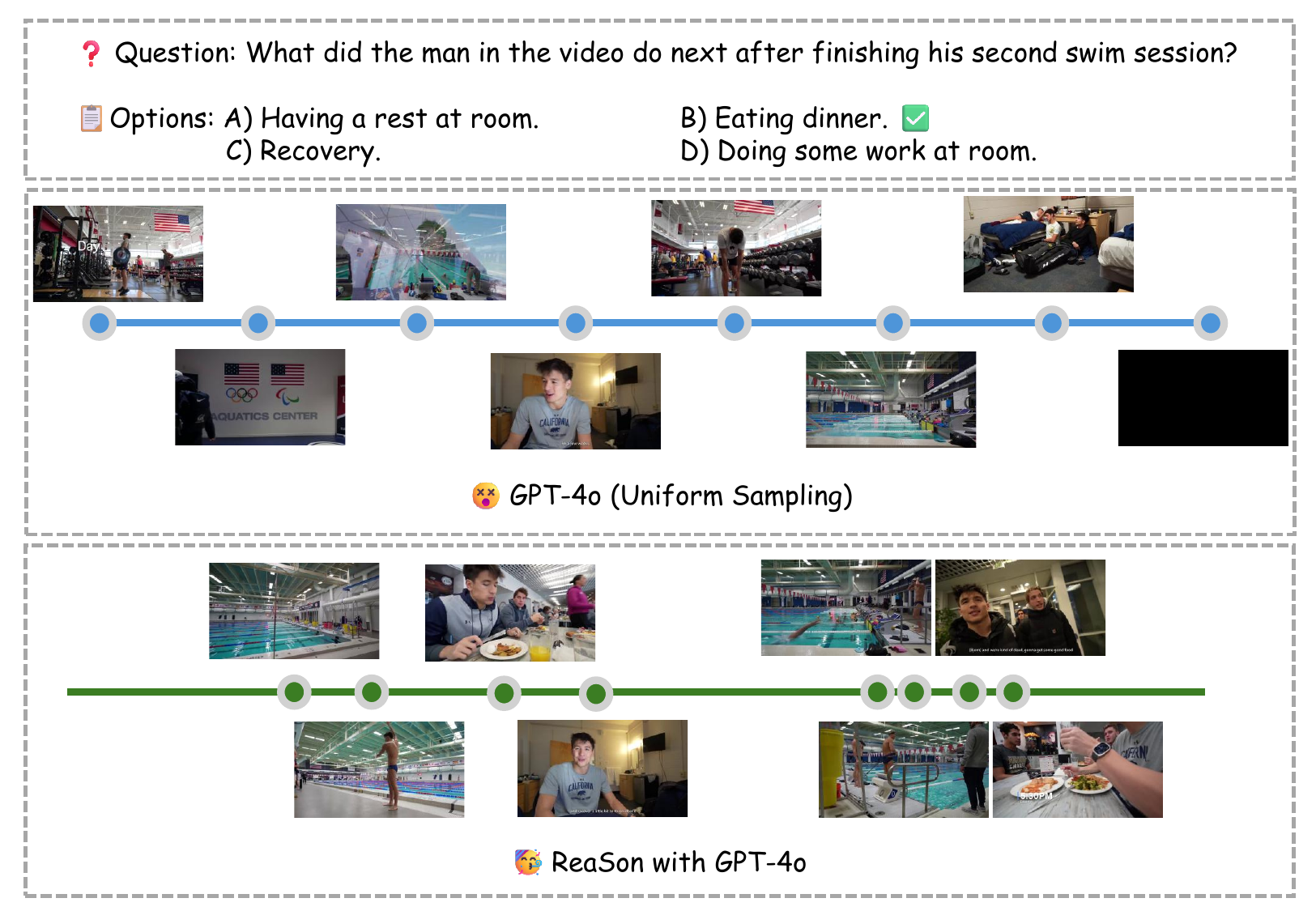}
    \caption{A comparative case study between GPT-4o with uniform sampling and \textbf{ReaSon}-enhanced frame selection. The example is drawn from a 9-minute video in Video-MME under the 8-frame setting. GPT-4o fails to capture the relevant scene with uniformly sampled frames, while ReaSon selects relevant and decisive frames that lead to the correct answer.}
    \label{casestudy}
\end{figure*}

\begin{figure*}[htbp]
\centering
    \includegraphics[width=0.9\linewidth]{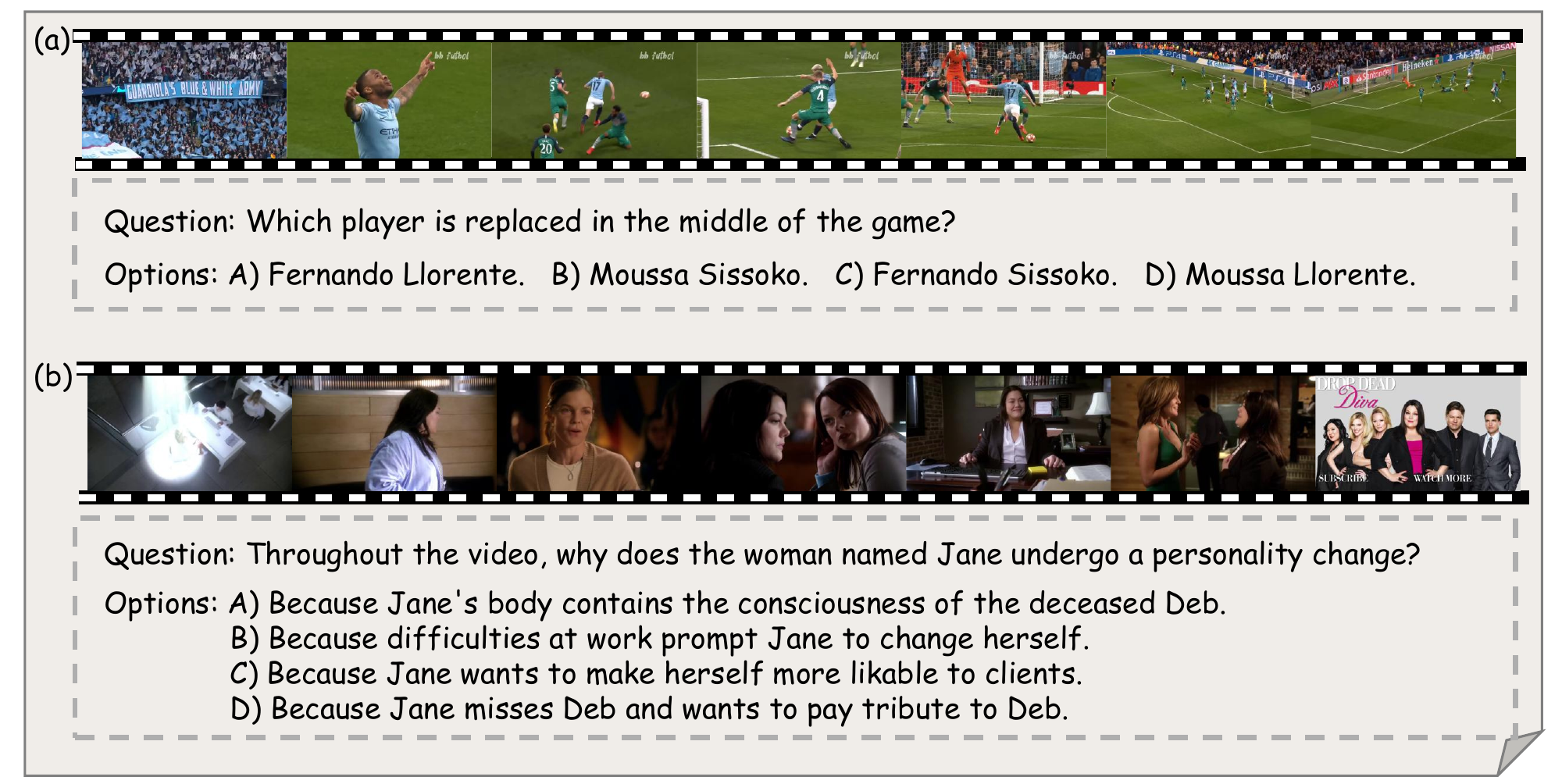}
    \caption{Limitation cases sampled from long videos in Video-MME. (a) A question requiring external knowledge to identify a specific athlete. (b) A question requiring long-range tracking of character states in episodic content.}
    \label{limitationcase}
\end{figure*}

\section{Case Study}
Fig.~\ref{casestudy} shows a comparative case from a 9-minute video in the Video-MME dataset, where the question asks, “What did the man do next after finishing his second swim session?” Under the 8-frame constraint, GPT-4o with uniform sampling receives temporally scattered frames, many of which fail to capture the relevant post-swim activity. As a result, it misinterprets the temporal context. 

In contrast, ReaSon selects a compact subset of frames centered around the second swim session and the subsequent dining scene. Our method provides GPT-4o with contextually relevant and temporally aligned inputs, enabling it to produce the correct answer, “Eating dinner.” This example highlights the limitation of uniform sampling in long videos, where crucial information may be sparsely located, and demonstrates the benefit of ReaSon in capturing decisive moments necessary for accurate temporal reasoning.

\section{Limitations}
While ReaSon demonstrates strong performance under limited-frame budgets, we outline two practical scenarios beyond its current design scope, as shown in Fig.~\ref{limitationcase}. \emph{(a) External-knowledge requirement.} Questions that rely on background facts, such as identifying an athlete or a specific team in sports footage, require the combination of VLM with a dedicated knowledge-retrieval module, which lies outside the scope of our CIB objective. \emph{(b) Long-range entity tracking.} In long-form or episodic videos, consistent entity tracking is essential for stitching together related events over extended horizons. Our method is designed to identify sufficient and necessary keyframes, enabling reliable reasoning with limited frames and computational overhead. Incorporating VLM-based external-knowledge retrieval and long-range tracking mechanisms is left to future work.

\end{document}